\pdfoutput=1

\documentclass[11pt]{article}

\usepackage[]{acl}

\usepackage{times}
\usepackage{latexsym}

\usepackage[T1]{fontenc}

\usepackage[utf8]{inputenc}

\usepackage{microtype}
\usepackage{inconsolata}
\usepackage{graphicx}
\usepackage{hyperref}
\usepackage{lipsum}
\usepackage{url}

\title{EaSyGuide : ESG Issue Identification Framework leveraging Abilities of Generative Large Language Models}

\author{
  Hanwool Lee, Jonghyun Choi, Sohyeon Kwon, Sungbum Jung \\
  NCSOFT \\
  \texttt{\{albertmade, excelsiorcjh, sohyeonk, successtiger\}@ncsoft.com}
}
        
\begin{document}
\maketitle
\begin{abstract}
This paper presents our participation in the FinNLP-2023 shared task on multi-lingual environmental, social, and corporate governance issue identification (ML-ESG). The task's objective is to classify news articles based on the 35 ESG key issues defined by the MSCI ESG rating guidelines. Our approach focuses on the English and French subtasks, employing the CerebrasGPT, OPT, and Pythia models, along with the zero-shot and GPT3Mix Augmentation techniques. We utilize various encoder models, such as RoBERTa, DeBERTa, and FinBERT, subjecting them to knowledge distillation and additional training.

Our approach yielded exceptional results, securing the first position in the English text subtask with F1-score 0.69 and the second position in the French text subtask with F1-score 0.78. These outcomes underscore the effectiveness of our methodology in identifying ESG issues in news articles across different languages. Our findings contribute to the exploration of ESG topics and highlight the potential of leveraging advanced language models for ESG issue identification.
\end{abstract}

\section{Introduction}

Environmental, Social, and Governance (ESG) factors have gained significant attention in the realm of corporate sustainability in recent years. Companies are increasingly recognizing the profound impact of ESG practices on their long-term success and resilience. Numerous research have highlighted the positive correlation between robust ESG strategies and improved financial performance\cite{eccles2019social}. For instance, a comprehensive meta-analysis of over 2000 empirical studies revealed a positive correlation between ESG and corporate financial performance, indicating the integral role of ESG in value creation\cite{friede2015social}. Consequently, understanding and integrating ESG principles into corporate strategies have become crucial for ensuring sustainable and resilient businesses in the modern era.

In parallel, there has been a growing recognition of the importance of leveraging natural language processing (NLP) techniques to incorporate ESG factors effectively. The integration of NLP holds great potential for enhancing our understanding of ESG-related information and its impact on businesses and society. By leveraging NLP, we can effectively analyze and extract insights from vast amounts of textual data, such as news articles, to gain deeper insights into companies' ESG performance and their societal impact.

Motivated by these developments, our team participated in the FinNLP-2023 shared task on multi-lingual ESG issue identification(ML-ESG)\cite{chen2023ESG}. The objective was to classify ESG-related news articles into 35 key issues based on the MSCI ESG rating guidelines. To accomplish this, We employed useful techniques such as Zero-shot and GPT3Mix Augmentation. Furthermore, we trained and evaluated various encoder models to assess their performance in the English and French text domains. Our best-performing model ranked first in the English Text subtask and second in the French Text subtask, highlighting the effectiveness of our approach in advancing NLP capabilities for ESG issue identification.

\section{SharedTask ML-ESG}

The SharedTask ML-ESG focuses on identifying ESG issues in news articles written in multiple languages. It builds upon the FinSim4-ESG shared task\cite{kang-el-maarouf-2022-finsim4} from FinNLP-2022. Our participation was specifically in the English subtask and the French task. The goal is to classify news articles into 35 ESG key issues based on MSCI ESG rating guidelines. The dataset includes separate training and testing sets in English and French, with 1,119 English articles and 1,200 French articles in the training set, and 300 articles in each language in the testing set.

\section{Approaches}

\begin{figure}[t!]
    \centering
    \includegraphics[width=\columnwidth]{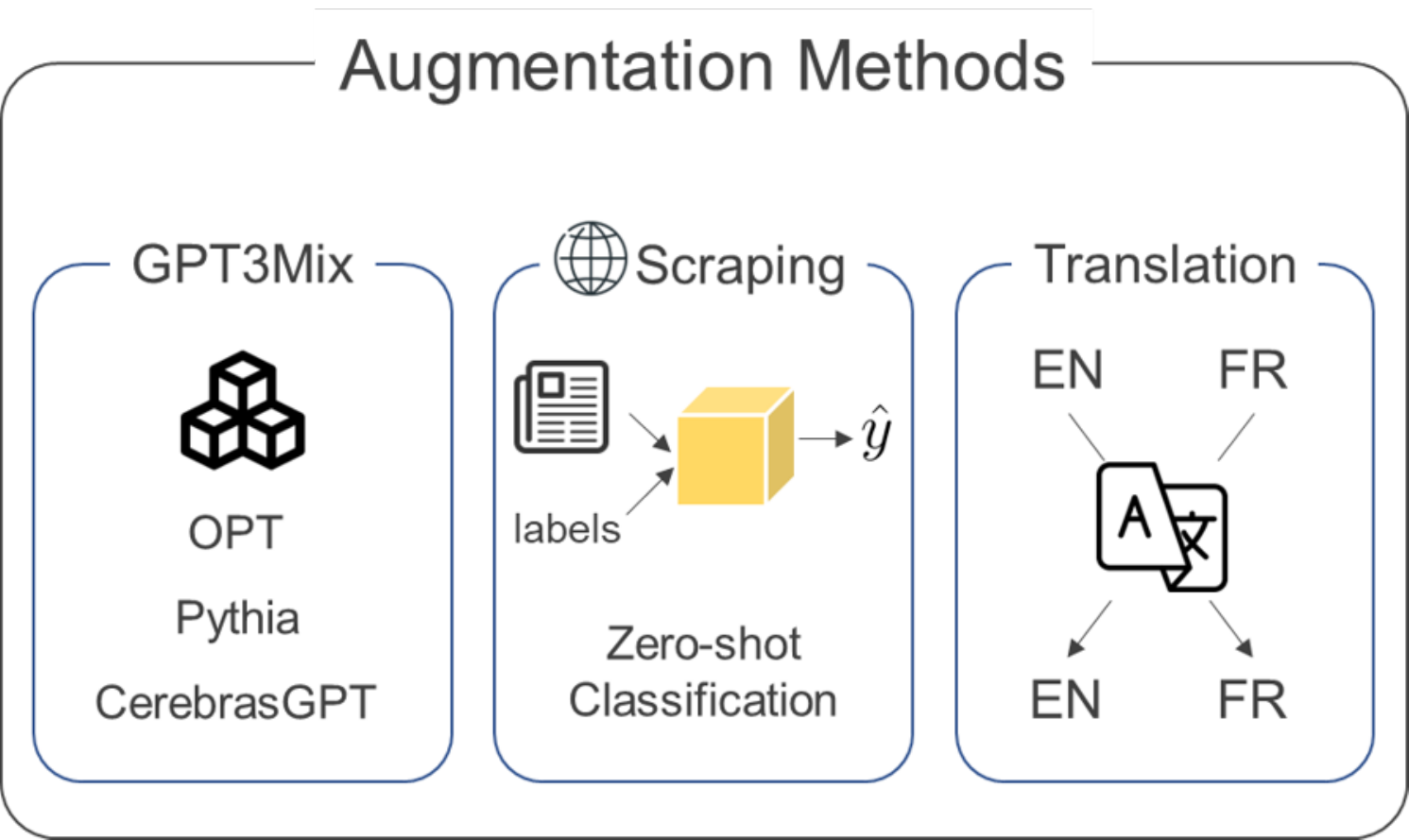}
    \caption{Overview of our approaches}
    \label{experiment_}
\end{figure}

The primary objective of our work is to distill the capabilities of various well-known generative large language models to create a lightweight yet powerful encoder model. ESG Issue classification allows for generative model and sequence-to-sequence model approaches, but due to limitations in our computing resources and time constraints, we devised an approach focused primarily on the encoder models. Given the ML-ESG task's limited sample size (around 1200) and imbalanced label distribution, training on the available data alone is insufficient to fully train on 35 labels. To overcome these challenges, we employed three renowned open-source generative models: Pythia\cite{biderman2023pythia}, CerebrasGPT\cite{dey2023cerebrasgpt}, and OPT\cite{zhang2022opt}. Due to limitations in computational resources, we utilized a 12B model for Pythia, while CerebrasGPT and OPT utilized 13B models.

\subsection{GPT3Mix}

To augment the available data, we employed the GPT3Mix\cite{yoo-etal-2021-gpt3mix-leveraging} technique, which leverages large-scale language models to generate synthetic text samples. By blending real samples and leveraging soft-labels derived from the language models, GPT3Mix captures the intricacies of human language effectively. We integrated the MSCI guideline's label descriptions into the GPT3Mix template, enhancing the generation process and ensuring augmented data aligns with the desired label semantics.

\subsection{Zero-Shot Classification}

The ML-ESG task includes English and French subtasks, each with 35 classification criteria based on the MSCI guideline. However, the complexity of each criterion's decision boundaries poses challenges when relying solely on the available training data. To address this, we performed zero-shot classification\cite{xian2020zeroshot} using ESG-related news collected through web scraping. To prevent prior exposure to the model, we excluded news articles from the training set originating from the same sources as the train set. We ensured label consistency by utilizing additional data only when Pythia, CerebrasGPT, and OPT provided identical labeling.

\subsection{Translation}

To train both multilingual and monolingual models, we leveraged translated versions of the English and French training sets as additional data. For translation, we utilized the widely recognized translation service, Deepl~\footnote{\url{https://www.deepl.com/translator}}.

By employing these approaches, we aimed to optimize the use of generative models, apply data augmentation through GPT3Mix, perform zero-shot classification, and incorporate translated data to enhance the capabilities of our encoder models for the ML-ESG task.

\section{Experiments and Results}
Our experiments were conducted in two phases. In the first phase, our aim was to identify effective encoder models and approaches by combining various techniques. In the second phase, we aimed to build an optimal model based on the findings from the first phase.

All experiments were conducted using the same hyperparameters: learning rate of 3e-4, epoch of 20, and optimizer of AdamW\cite{loshchilov2019decoupled}. The experiments were run on a single A100 GPU.

\subsection{First Experiment}

\begin{table}[ht]
\centering
\begin{tabular}{lccc}
\hline
\textbf{Model} & \textbf{Subtask} & \textbf{Valid F1} & \textbf{Test F1} \\
\hline
RoBERTa-base & English & 0.66 & 0.67 \\
DeBERTa-large & English & 0.65 & 0.69 \\
FinBERT & English & 0.53 & 0.56 \\
mRoBERTa-xl & English & 0.61 & 0.69 \\
DeBERTa-base & English & 0.51 & 0.58 \\
mDeBERTa & English & 0.44 & 0.52 \\
mRoBERTa-xl & French & 0.76 & 0.75 \\
mDeBERTa & French & 0.49 & 0.47 \\
\hline
\end{tabular}
\caption{Overview of baseline experiment results}
\label{baseline}
\end{table}

In the first experiment, we aimed to validate the performance of various encoder models for the ML-ESG task. We utilized well-known encoder models, including DeBERTa\cite{he2021deberta}, RoBERTa\cite{liu2019roberta}, and FinBERT\cite{araci2019finbert} which is specifically designed for financial text. To ensure applicability across both English and French subtasks, we also incorporated multilingual encoder models, namely mDeBERTa\cite{he2023debertav3} and mRoBERTa\cite{goyal2021largerscale}.

To evaluate the capabilities of these models, we employed stratified sampling to extract a validation set comprising approximately 5\% of the training set. Due to the unbalanced label distribution, we utilized the weighted F1 score as the primary evaluation metric. Baseline scores were obtained for each model, and any model with a validation F1 score below 0.45 was excluded from further experimentation. The summarized performance results of the baseline models are presented in the table ~\ref{baseline}, serving as the baseline for further experiments.

\begin{table}[ht]
\centering
\resizebox{\columnwidth}{!}{%
\begin{tabular}{lcc}
\hline
\textbf{Method} & \textbf{EN} & \textbf{FR} \\
\hline
Original & 1199 & 1200 \\
GPTMix-OPT (opt) & 2866 & 2867 \\
GPTMix-Pythia (pyt) & 2900 & 2901 \\
GPTMix-CerebrasGPT (cpt) & 2906 & 2907 \\
GPTMix-Mixed Models (mix) & 7473 & 7474 \\
Crawled (da) & 4816 & - \\
Translation (ts) & 2279 & 2279 \\
\hline
\end{tabular}%
}
\caption{Size of dataset for each approaches, The content within parentheses represents the abbreviation of the respective datasets.}
\label{data-size}
\end{table}

In addition to the initial experimentation, we employed data augmentation techniques to further enhance the performance of our models. We leveraged large-scale language models, including Pythia, CerebrasGPT, and OPT, for augmentation. For each news article, we generated additional samples and removed poorly generated ones to form a training dataset for each GPT3Mix augmentation technique.

Furthermore, we crawled ESG-related news articles in both languages, assigning labels to the collected data using Zero-shot Classification. Duplicate labels were removed, resulting in a cleaner dataset. Additionally, we added a translated version of the original training dataset to train monolingual models for English and French. We also constructed the 'GPTMix-Mixed Models (mix)' dataset by aggregating all GPT3Mix Augmentation datasets for further experimentation. Finally, we merged the augmented data for English and French to train a multilingual model.
\begin{table}[ht]
\centering
\resizebox{\columnwidth}{!}{%
\begin{tabular}{lccc}
\hline
\textbf{Experiment Name} & \textbf{Subtask} & \textbf{Valid F1} & \textbf{Test F1} \\
\hline
RoBERTa-base-mix & English & 0.749 & 0.597 \\
DeBERTa-large-ts & English & 0.737 & 0.705 \\
RoBERTa-base-pyt & English & 0.735 & 0.629 \\
RoBERTa-base-opt & English & 0.730 & 0.603 \\
RoBERTa-base-cpt & English & 0.709 & 0.628 \\
DeBERTa-base-da & English & 0.694 & 0.615 \\
mDeBERTa-mix & French & 0.760 & 0.731 \\
mRoBERTa-xl-cpt & French & 0.702 & 0.714 \\
mDeBERTa-pyt & French & 0.671 & 0.663 \\
mDeBERTa-opt & French & 0.657 & 0.656 \\
mRoBERTa-xl-ts & French & 0.625 & 0.695 \\
\hline
\end{tabular}%
}
\caption{Best performing models for each methodology}
\label{best-performing}
\end{table}

Among the English models, "roberta-base-mix" trained on data augmented by large language models OPT, Pythia, and CerebrasGPT and subsequently merged, achieved the highest validation F1 score of 0.7489. Furthermore, models trained on data augmented through translation and crawling obtained higher validation F1 scores than those trained on the original dataset. These results demonstrate the significant effectiveness of our proposed augmentation methodologies.

Similarly, for the French subtask, "mdeberta-mix" trained on data augmented using large language models, achieved a high validation F1 score of 0.7602, indicating that most of models trained on augmented data outperformed the baselines.

These experimental results highlight the efficacy of our approach and the positive impact of data augmentation on the performance of the encoder models in the ML-ESG task.

\subsection{Second Experiment}
Experiment 1 aimed to analyze the performance of various encoder models in multifarious ways. In contrast, experiment 2 focused on conducting experiments on several datasets using both base and large models from a specific subset of models, providing a more targeted investigation.

\begin{figure}[t!]
    \centering
    \includegraphics[width=\columnwidth]{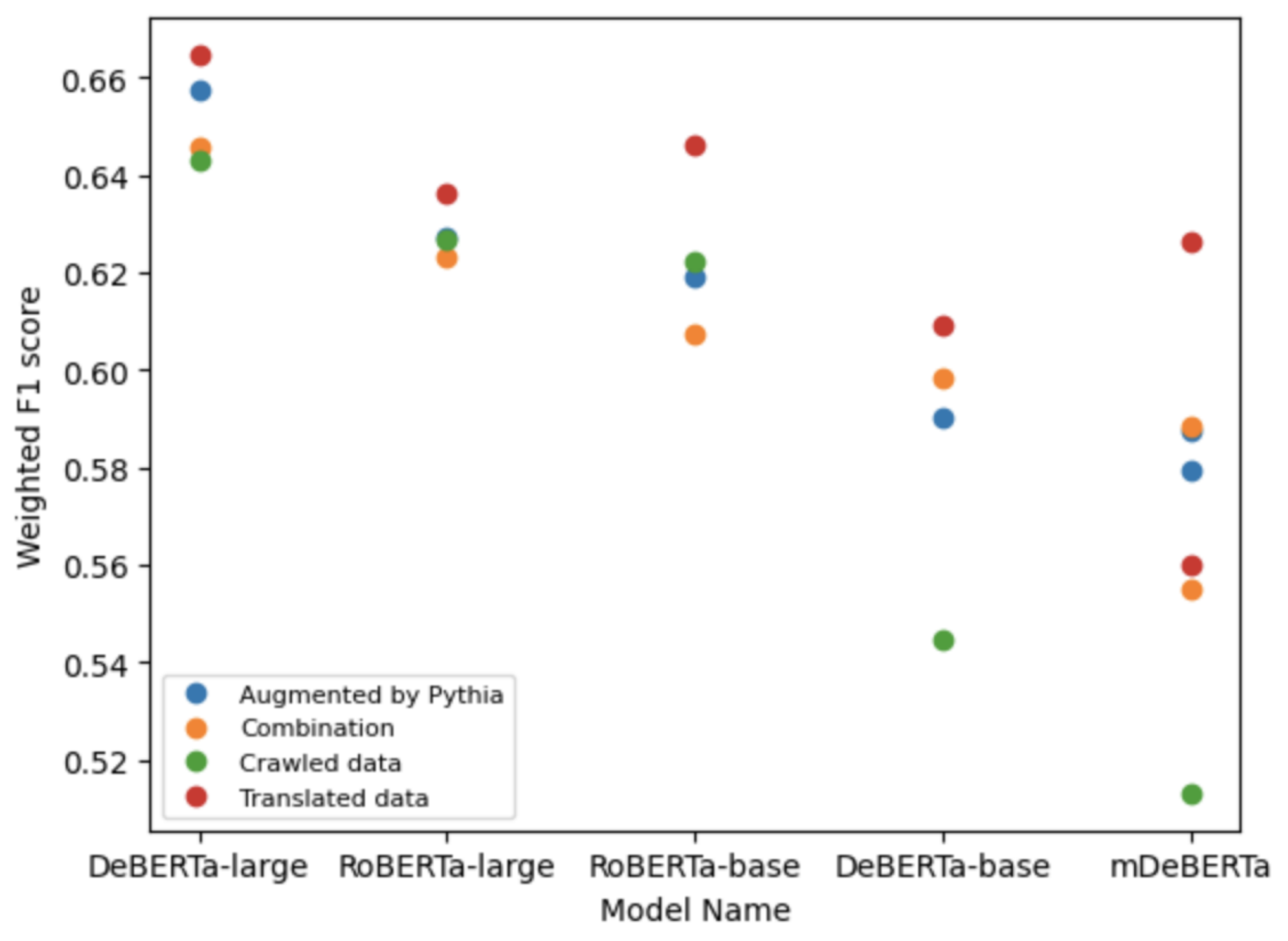}
    \caption{Experiment Results for experiment 2 on test set}
    \label{experiment2}
\end{figure}

\subsubsection{Effect of Model Size}
The size of language models is a significant factor that impacts their performance. In this experiment, we compared the classification performance of two well-performing language models from Experiment 1, DeBERTa and RoBERTa, at their base and large versions to analyze the effect of model size.

Figure~\ref{experiment2} illustrates that the large models of RoBERTa and DeBERTa consistently achieved higher F1 scores compared to their base models. Notably, the F1 scores of the large models were concentrated within a narrower range, indicating a more reliable and accurate performance. This suggests that the base models do not offer superior performance and are more susceptible to bias in classification results, struggling to accurately distinguish certain labels.

\subsubsection{Model Robustness}
Despite our efforts to construct a training dataset with a balanced label distribution in Experiment 1, our language models faced challenges in handling out-of-distribution issues. The complexity arose from the large number of MSCI ESG standard labels (35 in total), which strained the models' predictive abilities. Certain labels, such as controversial sourcing, revealed noticeable weaknesses in our models' predictions.

To address these shortcomings and enhance the robustness of the language models, we modified our training and validation datasets in the second experiment. We allocated a greater number of samples to labels from web-crawled dataset that had proven challenging for the models to predict accurately.

\subsubsection{Integration of data and ensembling}
Our research aimed to enhance ESG issue classification in a multilingual context by adopting a diverse and multi-faceted approach. We utilized four types of datasets(ts,pyt,da,combined) and experimented with eight different models, exploring data mixing and ensemble methods to optimize model performance.

Although the combined datasets showed promising performance, they did not outperform models trained exclusively on translated data in second experiment. This indicates that incorporating data from diverse sources may introduce additional noise and potentially decrease performance. In such circumstances, ensembling the results from various models proved beneficial\cite{RUTA200563}. For the English task, we employed a hard-voting ensemble of the top-scoring models, trained on different datasets using various encoder models. This ensemble approach achieved the highest performance, with an F1 score of 0.69 on the test set and 0.81 on the validation set, demonstrating the effectiveness of combining diverse models and datasets. Similarly, for the French task, we applied an ensemble technique by combining predictions from three models trained on different datasets and diverse encoder models, resulting in an impressive F1 score of 0.78 on the test set (0.8 on the validation set), further highlighting the effectiveness of combining models in a multilingual context for ESG issue classification.

\section{Conclusion}
In this paper, we presented our approach for the FinNLP-2023 shared task on multi-lingual ESG issue identification. By leveraging advanced encoder models and techniques like GPT3Mix Augmentation, zero-shot classification, and translation, we achieved promising results. Our models ranked first in the English text subtask and second in the French text subtask, highlighting the effectiveness of our methodology across different languages. Our research contributes to exploring ESG topics and showcases the potential of advanced language models in identifying ESG issues. Future work would focus on exploring decoder and sequence-to-sequence architectures, expanding to other languages, and employing alternative models to improve the accuracy and generalizability of ESG issue identification systems.

\section*{Availability}

The code is available at \url{https://github.com/finMU/ML-ESG_codes}.

\bibliography{anthology,custom}

\appendix

\section{Appendix}
\subsection{GPT3Mix Prompt Details}
Based on the GPT3Mix paper, we developed a Task Specification Template for the MLESG Shared Task and randomly extracted examples. However, considering the high probability of introducing imbalanced data when extracting augmented data with imbalanced labels, we equalized the extraction probability for each label to mitigate the imbalance issue. In this process, including descriptions for all 35 labels in the prompt could lead to excessive context, so we only utilized label descriptions for the labels present in the samples. Below is an example showcasing a partial portion of the prompt we employed.

\subsubsection{Example of Task Description}
\begin{verbatim}
Each item in the following list should contain
#ESG News headline, #ESG news and the related 
#ESG key issues. #ESG key issues are based on 
MSCI ESG rating guidelines.
\end{verbatim}

\subsubsection{Example of Label Description}
\begin{verbatim}
Access to Finance: This label is about their 
efforts to expand financial services to 
historically underserved markets, including 
small-business lending and the development of 
innovative distribution channels.
\end{verbatim}

\subsection{Data distributions}
In our study, we utilized GPT3Mix to augment the dataset, resulting in a well-balanced distribution of labels. Each of the 35 labels accounted for approximately 2.85\% (plus or minus 0.7\%) of the dataset. However, when performing zero-shot classification on the data obtained through web crawling, we encountered limitations. This was due to either the scarcity of relevant data available on the web or the presence of insufficient labels caused by model bias. To provide further insights, we present a detailed table~\ref{label-distribution} showcasing the label distribution exclusively based on the crawled data.

\begin{table}[h]
\centering
\resizebox{0.5\textwidth}{!}{%
\begin{tabular}{p{0.70\linewidth}p{0.30\linewidth}}
\hline
\textbf{Label} & \textbf{Percentage (\%)} \\
\hline
Board & 8.80 \\
Carbon Emissions & 6.94 \\
Responsible Investment & 5.64 \\
Accounting & 5.53 \\
Pay & 5.36 \\
Packaging Material \& Waste & 4.57 \\
Business Ethics & 4.35 \\
Water Stress & 4.23 \\
Financing Environmental Impact & 4.01 \\
Opportunities in Renewable Energy & 3.89 \\
Human Capital Development & 3.84 \\
Community Relations & 3.78 \\
Consumer Financial Protection & 3.78 \\
Product Carbon Footprint & 3.67 \\
Opportunities in Clean Tech & 3.50 \\
Biodiversity \& Land Use & 2.99 \\
Electronic Waste & 2.82 \\
Chemical Safety & 2.60 \\
Raw Material Sourcing & 2.54 \\
Opportunities in Green Building & 2.48 \\
Ownership \& Control & 2.37 \\
Climate Change Vulnerability & 1.81 \\
Toxic Emissions \& Waste & 1.58 \\
Health \& Demographic Risk & 1.24 \\
Access to Finance & 1.24 \\
Opportunities in Nutrition \& Health & 1.24 \\
Access to Health Care & 1.19 \\
Privacy \& Data Security & 0.96 \\
Access to Communications & 0.73 \\
Product Safety \& Quality & 0.68 \\
Supply Chain Labor Standards & 0.62 \\
Labor Management & 0.51 \\
Controversial Sourcing & 0.51 \\
\hline
\end{tabular}%
}
\caption{Label Distribution of Crawled Data}
\label{label-distribution}
\end{table}

\end{document}